# A Study of Generative Large Language Model for Medical Research and Healthcare

**Authors:** Cheng Peng[1], Xi Yang[1,2][†], Aokun Chen[1,2], Kaleb E Smith[3], Nima PourNejatian[3], Anthony B Costa[3], Cheryl Martin[3], Mona G Flores[3], Ying Zhang[4], Tanja Magoc[5], Gloria Lipori[5,6], Duane A Mitchell[6], Naykky S Ospina[7], Mustafa M Ahmed[8], William R Hogan[1], Elizabeth A Shenkman[1], Yi Guo[1,2], Jiang Bian[1,2], Yonghui Wu[1,2] *

**Affiliations:**

[1]Department of Health Outcomes and Biomedical Informatics, College of Medicine, University of Florida, Gainesville, Florida, USA.

[2]Cancer Informatics Shared Resource, University of Florida Health Cancer Center, Gainesville, Florida, USA.

[3]NVIDIA, Santa Clara, California, USA.

[4]Research Computing, University of Florida, Gainesville, Florida, USA.

[5]Integrated Data Repository Research Services, University of Florida, Gainesville, Florida, USA.

[6]Lillian S. Wells Department of Neurosurgery, UF Clinical and Translational Science Institute, University of Florida.

[7]Division of Endocrinology, [8]Division of Cardiovascular Medicine, Department of Medicine, College of Medicine, University of Florida, Gainesville, FL, USA

[†]Xi Yang finished this work when he was a full-time employee at the University of Florida.

*Corresponding author

Yonghui Wu, PhD

Clinical and Translational Research Building

2004 Mowry Road, PO Box 100177, Gainesville, FL, USA, 32610

Phone: 352-294-8436

Email: yonghui.wu@ufl.edu

Word count : 4000 max

**Abstract**

There is enormous enthusiasm and concerns in using large language models (LLMs) in healthcare, yet current assumptions are all based on general-purpose LLMs such as ChatGPT. This study develops a clinical generative LLM, GatorTronGPT, using 277 billion words of mixed clinical and English text with a GPT-3 architecture of 20 billion parameters. GatorTronGPT improves biomedical natural language processing for medical research. Synthetic NLP models trained using GatorTronGPT generated text outperform NLP models trained using real-world clinical text. Physicians' Turing test using 1 (worst) to 9 (best) scale shows that there is no significant difference in linguistic readability ($p = 0.22$; 6.57 of GatorTronGPT compared with 6.93 of human) and clinical relevance ($p = 0.91$; 7.0 of GatorTronGPT compared with 6.97 of human) and that physicians cannot differentiate them ($p < 0.001$). This study provides insights on the opportunities and challenges of LLMs for medical research and healthcare.

Generative large language models (LLMs) such as the ChatGPT[1] have surprised the world by answering questions conversationally and generating decent textual contents such as emails, articles, and even computer codes, triggering enormous enthusiasm in potential applications for medical research and healthcare.[2–4] People are enthusiastic about the potential of using LLMs to facilitate documentation of patient reports (e.g., a progress report),[3,4] improving diagnostic accuracy,[5] and assisting in various clinical care,[6,7] while at the same time concerning about the hallucinations and fabrications,[7,8] bias and stereotype,[9] and risks of patient privacy and ethics.[10] Yet, this enthusiasm and concerns are based on a general-purpose LLM ChatGPT, which is not designed for healthcare use since only a small fraction of biomedical text was used.[1] Until now, it is unclear how this disruptive technology can help medical research and potentially improve the quality of healthcare.

Language model is a simple statistical distribution used in natural language processing (NLP) to formulate the probability of a sequence of words or the next word in a sequence. Surprisingly, when it is used as a self-supervised learning objective to train a specific neural network architecture named transformer, and when the model size is very large such as billions or hundreds of billions of parameters, important artificial intelligence (AI) emerge. For example, LLMs can learn knowledge from one task and apply it to another task (i.e., transfer learning), learn from very few labeled samples (i.e., few-shot learning), and learn without human labeled samples for the target application (i.e., zero-shot learning).[11–13] The pretrained transformer architecture is known as generative LLM as it can generate human-like text. The conversational ability of LLMs is achieved using prompt-based text generation,[14] the key technology guiding LLMs to generate reasonable answers and contextual contents.

This study aims to develop a generative LLM in the medical domain and evaluate its utility for medical research and healthcare. We trained a generative LLM, namely GatorTronGPT, using 82 billion words of de-identified clinical text[15] from University of Florida (UF) Health and 195 billion diverse English words from the Pile[16] dataset. We trained GatorTronGPT from scratch using the GPT-3[17] architecture (used by ChatGPT) and examined how the text generation ability of GatorTronGPT benefit medical research and healthcare. We formulated biomedical relation extraction and question answering using a unified text generation architecture[18] to evaluate how GatorTronGPT could benefit medical research using 6 benchmark datasets. To examine the utility of text generation in the clinical domain, we applied GatorTronGPT to generate 20 billion words of synthetic clinical text, which were used to train synthetic NLP models, denoted as GatorTronS ('S' stands for synthetic). We compared GatorTronS models with GatorTron,[15] a clinical NLP model trained with the same architecture but using real-world 90 billion words of text, on 5 different clinical NLP tasks to test the hypothesis that generative clinical LLMs can be used to generate synthetic clinical texts useful for clinical research. To test if LLMs could be used in healthcare, two internal medicine subspecialists from endocrinology (NSO) and cardiology (MMA) manually evaluated 60 clinical paragraphs including 30 paragraphs written by GatorTronGPT randomly mixed with 30 real-world paragraphs written by UF Health physicians. **Fig. 1** shows an overview of the study design. To our best knowledge, GatorTronGPT is the first generative LLM developed in the clinical domain using the GPT-3 architecture with 20 billion parameters, providing valuable insights on the opportunities and challenges of generative LLMs for medical research and healthcare.

**Results**

We trained GatorTronGPT using 5 billion and 20 billion parameters with 277 billion words of mixed clinical and general English text. Training the 5 billion model used approximately 6 days and the 20 billion model used about 20 days on 560 A100 80G GPUs from 70 NVIDIA DGX notes using the NVIDIA SuperPOD reference cluster architecture. **Fig. 2** shows the training and validation loss for the two sizes of GatorTronGPT models.

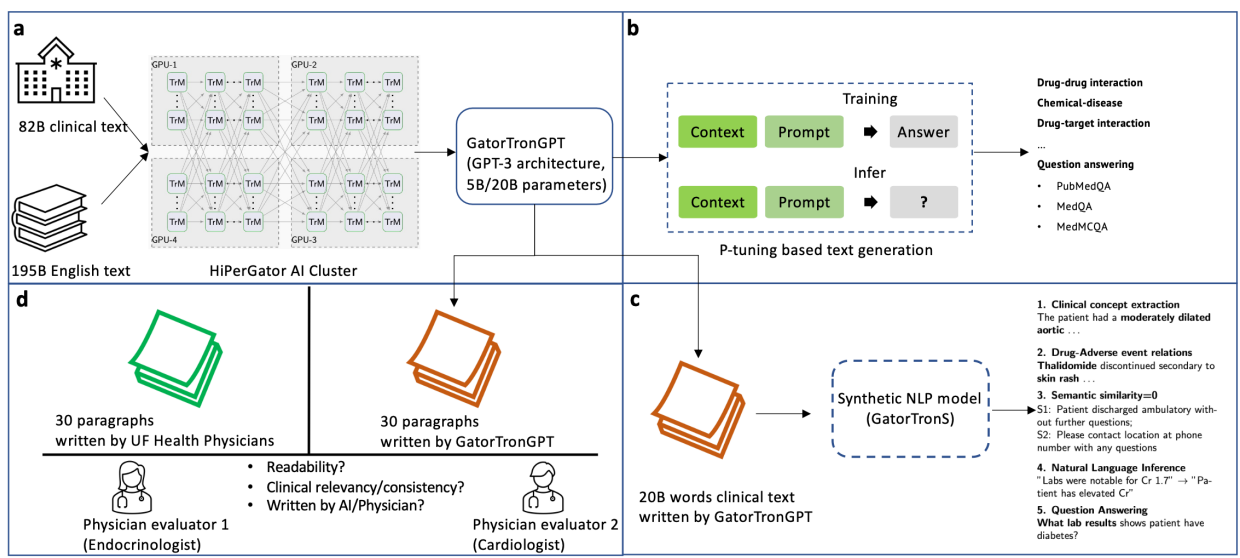

**Fig 1. Develop a clinical generative large language model, GatorTronGPT, for biomedical natural language processing, clinical text generation, and healthcare text evaluation. a**, Train GatorTronGPT from scratch using GPT-3 architecture with up to 20 billion parameters. **b,** Solve biomedical relation extraction and question answering using a unified P-tuning base text generation architecture. **c,** Apply GatorTronGPT to generate 20 billion words of synthetic clinical text, which was used to train synthetic natural language processing model, GatorTronS. **d,** Turing evaluation of 30 paragraphs of text written by GatorTronGPT mixed with 30 real-world paragraphs written by UF Health physicians. TrM: transformer unit; B: billion

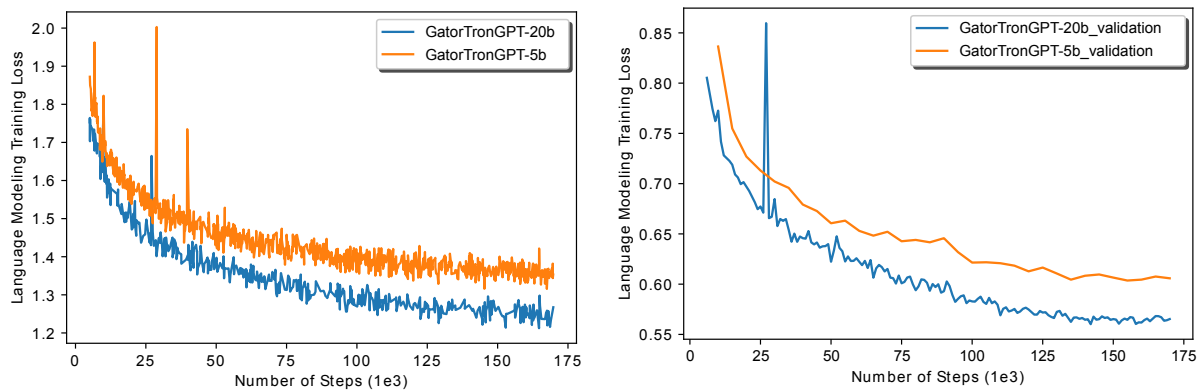

**Fig. 2** Training loss and validation loss for GatorTronGPT 5 billion and 20 billion models.

**Table 1.** Comparison of GatorTronGPT with existing transformer models for **a.** biomedical relation extraction and **b.** question answering.

**a**

|  | Biomedical Relation extraction | | | | | | | | |
|---|---|---|---|---|---|---|---|---|---|
|  | DDI | | | BC5CDR | | | KD-DTI | | |
| Model | Pre | Rec | F1 | Pre | Rec | F1 | Pre | Rec | F1 |
| GPT-2 medium | 0.234 | 0.319 | 0.247 | 0.439 | 0.326 | 0.374 | 0.305 | 0.279 | 0.285 |
| REBEL | 0.354 | 0.286 | 0.283 | 0.343 | 0.395 | 0.367 | 0.324 | 0.296 | 0.304 |
| REBEL-pt | 0.465 | 0.396 | 0.406 | 0.409 | 0.212 | 0.279 | 0.357 | 0.326 | 0.333 |
| BioGPT | 0.417 | 0.448 | 0.408 | 0.494 | 0.412 | 0.450 | 0.400 | 0.397 | 0.384 |
| GatorTronGPT-5B | 0.466 | 0.518 | 0.491 | **0.587** | 0.434 | 0.472 | 0.422 | 0.436 | 0.412 |
| GatorTronGPT-20B | **0.476** | **0.521** | **0.500** | 0.543 | **0.499** | **0.494** | **0.422** | **0.440** | **0.419** |

DDI: drug-drug interaction; BC5CDR: BioCreative V chemical-disease relation; KD-DTI: drug-target interaction; B: billion parameters. The best evaluation scores are bolded.

**b**

|  | Question answering | | |
|---|---|---|---|
|  | PubMedQA | MedQA (USMLE) | MedMCQA |
| Model | Accuracy | Accuracy | Accuracy |
| PubMedBERT | 0.558 | 0.381 | NA |
| BioELECTRa | 0.642 | NA | NA |
| BioLinkBERT | 0.702 | **0.451** | NA |
| GPT-2 | 0.750 | 0.333 | NA |
| BioGPT | **0.782** | NA | NA |
| Galactica_120B | 0.776 | 0.444 | **0.529** |
| GatorTronGPT-5B | 0.758 | 0.402 | 0.358 |
| GatorTronGPT-20B | 0.776 | **0.451** | 0.429 |

NA: performance not reported; B: billion parameters. The best evaluation scores are bolded.

**Table 1.a** compares GatorTronGPT with four existing biomedical transformer models on end-to-end relation extraction of drug-drug interaction, chemical-disease relation, and drug-target interaction. GatorTronGPT outperformed all existing transformer models on 3 datasets, where the GatorTronGPT with 20 billion parameters achieved the best F1-score of 0.500, 0.494, and 0.419, respectively. GatorTronGPT improved state-of-the-art by 3%-10% compared with the second-best bioGPT[18] model. We consistently observed performance improvement when scaling up the size of GatorTronGPT. **Table 1.b** compares GatorTronGPT with six existing biomedical transformers using three benchmark datasets for biomedical question answering. The GatorTronGPT model with 20 billion parameters achieved the best performance of 0.451, as a tie with BioLinkBERT, for the MedQA dataset, and achieved the second-best performance of 0.776 for the PubMedQA dataset. The performance of GatorTronGPT on the MedMCQA dataset is lower than a much larger LLM Galactica with 120 billion parameters. We observed a monotonic performance improvement by scaling up the size of GatorTronGPT.

**Table 2.** Comparison of GatorTronS with existing transformer-based LLMs for clinical concept extraction and medical relation extraction.

|  | Clinical concept extraction | | | | | | | | | Medical relation extraction | | |
|---|---|---|---|---|---|---|---|---|---|---|---|---|
|  | 2010 i2b2[19] | | | 2012 i2b2[20] | | | 2018 n2c2[21] | | | 2018 n2c2[21] | | |
| **Transformer** | Precision | Recall | F1 score | Precision | Recall | F1 score | Precision | Recall | F1 score | Precision | Recall | F1 score |
| ClinicalBERT | NA | NA | 0.878 | NA | NA | 0.789 | 0.859 | 0.883 | 0.871 | 0.968 | 0.941 | 0.954 |
| GatorTron, 90B | 0.875 | 0.904 | 0.889 | 0.764 | 0.822 | 0.792 | 0.876 | 0.904 | 0.890 | 0.972 | 0.948 | 0.960 |
| GatorTronS, 1B | 0.874 | 0.907 | 0.890 | 0.753 | 0.812 | 0.781 | 0.871 | 0.892 | 0.882 | 0.971 | 0.945 | 0.958 |
| GatorTronS, 5B | 0.879 | 0.909 | 0.894 | 0.777 | 0.823 | 0.799 | **0.899** | 0.903 | **0.901** | 0.974 | 0.949 | **0.962** |
| GatorTronS, 10B | 0.882 | **0.911** | 0.896 | 0.765 | 0.823 | 0.793 | 0.887 | 0.904 | 0.895 | 0.974 | **0.950** | **0.962** |
| GatorTronS, 20B | **0.889** | **0.911** | **0.899** | **0.784** | **0.836** | **0.809** | 0.892 | **0.907** | 0.900 | **0.975** | 0.947 | 0.961 |

B: billion words of text; Clinical concepts in 2010 i2b2 and 2012 i2b2 challenges: problems, treatments, lab tests; clinical concepts in 2018 n2c2 challenge: drugs, adverse events, and drug-related attributes (e.g., dose). Medical relation in 2018 n2c2 challenge: drug induced adverse events; B: billion words of text. Best evaluation scores are bolded. NA: scores not reported.

**Table 3.** Comparison of GatorTronS with existing transformer-based LLMs for semantic textual similarity, natural language inference, and question answering.

|  | Semantic textual similarity | Natural language inference | Question answering | | | |
|---|---|---|---|---|---|---|
|  | 2019 n2c2[22] | MedNLI[23] | emrQA Medication[24] | | emrQA Relation[24] | |
| **Transformer** | Pearson correlation | Accuracy | F1 score | Exact Match | F1 score | Exact Match |
| ClinicalBERT | 0.879 | 0.827 | 0.691 | 0.241 | 0.931 | 0.853 |
| GatorTron, 90B | 0.881 | 0.867 | 0.718 | 0.298 | 0.954 | 0.903 |
| GatorTronS, 1B | 0.853 | 0.851 | 0.702 | 0.288 | 0.965 | 0.924 |
| GatorTronS, 5B | 0.888 | 0.882 | 0.726 | 0.305 | 0.968 | 0.926 |
| GatorTronS, 10B | 0.893 | 0.886 | **0.728** | **0.311** | 0.972 | **0.929** |
| GatorTronS, 20B | 0.898 | **0.880** | 0.726 | 0.307 | **0.973** | 0.927 |

B: billion words of text. The best evaluation scores are bolded.

We generated 20 billion words of synthetic clinical text using GatorTronGPT. **Tables 2** and **3** compare GatorTronS trained with different sizes of synthetic clinical text with ClinicalBERT and the original GatorTron,[15] our previously released clinical LLM trained using real-world clinical text. For clinical concept extraction, the GatorTronS trained using 20 billion synthetic clinical text achieved the best F1-score for two out of three benchmark datasets, and GatorTronS trained using five billion synthetic clinical text achieved the best F1-score for 1 (the 2018 n2c2 challenge) out of three benchmark datasets. GatorTronS outperformed the original GatorTron model by >1% F1-score on all three benchmark datasets. For medical relation extraction, the GatorTronS trained using 10 billion synthetic clinical text achieved the best F1-score of 0.962 on the 2018 n2c2 challenge benchmark dataset, which is comparable with the original GatorTron model (0.960). For semantic textual similarity and natural language inference, the GatorTronS trained using 20 billion synthetic clinical text achieved the best evaluation scores, outperforming the original GatorTron by >1%. For question answering, the GatorTronS trained using 10 billion synthetic clinical text achieved the best score for emrQA benchmark focusing on medications, and the exact match evaluation for relation; the GatorTronS trained using 20 billion synthetic clinical text achieved the best evaluation score in F1-score evaluation on the emrQA relation benchmark dataset. GatorTronS outperformed the original GatorTron model trained using real-world clinical text > 1%. The comparison of GatorTronS models trained using different size of synthetic clinical text shows that by generating a minimum of 5 billion synthetic clinical text, we can train a synthetic GatorTronS model with comparable performance to GatorTron, a same size and architecture transformer trained using 90 billion words of clinical mixed with general English text.

The Turing test results show that, on average, less than half (49.2%) of the clinical notes were identified correctly, including 36.7% of the synthetic notes and 61.7% of the human notes (**Table 4.a**). Among the 30 synthetic notes written by GatorTronGPT, 9 (30.0%) and 13 (43.4%) were correctly labeled as 'AI' by the two physicians, respectively. Among the 30 human notes written by physicians, 17 (56.7%) and 20 (66.7%) were correctly labeled as 'Human', respectively. Considering GatorTronGPT was considered as a human for more than 30% of the instances (the criteria from Turing test),[25] GatorTronGPT passed the Turing test (p < 0.001). **Table 4.b** summarizes the means and standard deviations of the linguistic readability and clinical relevance and consistency. Statistical tests show that there is no significant difference between notes written by GatorTronGPT and human physicians in both linguistic readability (p = 0.22) and clinical relevance and consistency (p = 0.91). **Table 4.c** shows two examples of clinical paragraphs written by GatorTronGPT. Percent agreement and interrater reliability were found to be good or excellent, as summarized in Supplement Tables S1 and S2.

| a. Percentage of notes correctly identified by human reviewers. | | | |
|---|---|---|---|
| | **Ground truth** | | |
| | **AI** (n=30) | **Human** (n=30) | **Total** (n=60) |
| **Physician 1** | 9 (30.0%) | 17 (56.7%) | 26 (43.3%) |
| **Physician 2** | 13 (43.3%) | 20 (66.7%) | 33 (55%) |
| **Overall** | 11 (36.7%) | 18.5 (61.7%) | 29.5 (49.2%) |
| **p-value** | < 0.001 | = 0.104 | < 0.001 |

| b. Ratings of readability and clinical relevance by synthetic and human notes. | | | |
|---|---|---|---|
| | **Ground truth** | | |
| | **AI** (n=30) | **Human** (n=30) | p-value |
| **Readability** | 6.57 (1.22) | 6.93 (1.09) | 0.22 |
| **Clinical relevance** | 7.00 (1.23) | 6.97 (1.07) | 0.91 |

c.

**Example 1 (Passed Turing test)**: HISTORY OF PRESENT ILLNESS: Ms. [**NAME**] is a [**AGE**]-year-old female with prior history of left breast cancer who is referred to the [**LOCATION**] Clinic to discuss surgical management of her left lung nodule which was picked up on CT scanning for surveillance. The patient's chest CT obtained at an outside institution showed an incidental finding of a left lower lobe pleural nodule, which she brought with her to clinic today for further review. Ms. [**NAME**] reports to be doing well, with no change in her baseline level of shortness of breath, but denies any difficulty breathing, cough or hemoptysis. She feels essentially back to baseline and plays golf one day a week. Otherwise, Ms. [**NAME **] has no other pulmonary complaints and has never smoked tobacco or inhaled second hand smoke. She does report several years of intermittent reflux, which worsened after eating rough foods. She underwent an upper GI study for this, which showed a small hiatal hernia as well as a moderate- sized sliding hiatal hernia. She subsequently underwent an abdominal CT for evaluation for mesenteric ischemia and was felt to have a dilated pancreatic duct. Due to suspected head pathology on imaging (although the final report is pending), she underwent an MRI, which confirmed no pancreatic pathology. Given this, Ms. [**NAME **] has been trying to avoid fatty and fried foods, which has improved her reflux. Ms. [**NAME **] does report recent onset of sleep apnea, for which she is using CPAP every night for the last month. She tolerates this well without complaints of dry mouth, gasping for air, or undue sleepiness. Ms. [**NAME **] was seen in our outpatient [**LOCATION**] clinic on [**DATE**] at which time it was determined that in order to obtain a tissue diagnosis of this left lower lobe lesion a VATS approach would be utilized. She presents today for further pre-operative planning. Currently, Ms. [**DATE**] denies any pulmonary symptoms, fevers, chills, cough or weight loss. PAST MEDICAL HISTORY: 1. CARDIAC RISK FACTORS: Hypertension, Hyperlipidemia 2. CARDIAC HISTORY: None 3. OTHER PAST MEDICAL HISTORY: Left lower lobe pulmonary nodule

> **Example 2 (Failed Turing test):** Assessment: ICD-9-CM 1. Pain in joint, pelvic region and thigh, unspecified laterality 719.45 fentaNYL (DURAGESIC) 75 MCG/HR HYDROcodone-acetaminophen (VICODIN HP) 5-300 MG Tablet 2. Migraine NOS/not Intrcbl 346.90 SUMAtriptan Succinate 3. Bone Pain Arthritis 333.90 Dexamethasone Sodium Phosphate 4. Bipolar disorder 296.80 traZODone (DESYREL) 100 MG Tablet prazosin (MINIPRESS) 2 MG capsule carvedilol (COREG) 3.125 MG tablet isosorbide mononitrate (IMDUR) 30 MG CR tablet Refer to Psychiatry clopidogrel (PLAVIX) 75 MG tablet SUMAtriptan Succinate 5. ASTHMA UNSPECIFIED 493.90 albuterol (PROAIR HFA;VENTOLIN HFA) 108 (90 BASE) MCG/ACT inhaler 6. Major depressive disorder, single episode, unspecified 296.20 DULoxetine (CYMBALTA) 60 MG capsule Refer to Psychiatry amitriptyline (ELAVIL) 25 MG tablet traZODone (DESYREL) 100 MG Tablet 7. POST-SURGICAL VARICOSE VEINS of LOWER EXTREMITIES 454.9 fentaNYL (DURAGESIC) 75 MCG/HR 8. Other and unspecified hyperlipidemia 272.4 simvastatin (ZOCOR) 40 MG tablet COMPREHENSIVE METABOLIC PANEL 9. PND (post-nasal drip) 784.91 loratadine (CLARITIN) 10 MG tablet 10. Bipolar I disorder, single manic episode, unspecified 296.00 clonazePAM (KlonoPIN) 1 MG tablet Refer to Psychiatry 11. Allergic rhinitis 477.9 loratadine (CLARITIN) 10 MG tablet 12. Grief reaction 309.0 traZODone (DESYREL) 100 MG Tablet 13. Encounter for long-term (current) use of other medications V58.69 methocarbamol (ROBAXIN) 750 MG tablet COMPREHENSIVE METABOLIC PANEL 14. GERD (gastroesophageal reflux disease) 530.81 lansoprazole (PRE

**Table 4.** Turing test results. **a.** Number and percentage of correctly identified notes; **b.** Means and standard deviations of the quality measures; **c.** Two examples of synthetic clinical text generated by GatorTronGPT. The text generation stops at maximum 512 tokens. Pass Turing test: both physicians labeled as 'Human'; Fail Turing Test: both physicians labeled as 'AI'.

**Discussion**

This study develops a generative clinical LLM, GatorTronGPT, using the GPT-3 architecture[13] with 277 billion words of clinical mixed with English text. We evaluate GatorTronGPT for medical research and healthcare focusing on the key function of text generation. GatorTronGPT achieves state-of-the-art performance for 4 out 6 biomedical NLP benchmark datasets, demonstrating the benefit for medical research. The experimental results show that GatorTronGPT can generate synthetic clinical text for developing of synthetic clinical NLP models (i.e., GatorTronS), which achieve better or comparable performance with NLP models trained using real-world clinical text, demonstrating the utility of synthetic clinical text generation for clinical research. The physicians' evaluation of synthetic clinical text show that GatorTronGPT can generate clinical contents with linguistic readability comparable to real-world clinical notes. This study provides valuable insights regarding the opportunities and challenges of generative LLMs for medical research and healthcare.

We discover an important utility of generative LLMs for synthetic clinical text generation. There has been a gap in accessing large-scale clinical text and sharing clinical NLP models due to the sensitive nature of clinical text and the fact that automatic de-identification systems cannot remove 100% protected health information (PHI). Our study shows that GatorTronS, a synthetic transformer model trained using 5 billion words of synthetic clinical text generated by GatorTronGPT, can achieve better or comparable performance on 5 clinical NLP tasks compared with GatorTron[15], a same-structure and size transformer model trained using a much larger real-world clinical text (90 billion words). Potential reasons may include (1) real-world clinical text has redundancies, and (2) GatorTronGPT generates more diverse synthetic clinical text. A previous study[26] has reported that by augmenting real-world clinical training data using additional human annotated synthetic text generated by a smaller generative LLM, GPT-2, NLP models can achieve better performance. Our study further demonstrates that, without additional human annotation and augmentation of training data, a larger clinical GPT-3 model can generate synthetic clinical text to train synthetic NLP models outperforming NLP models trained using real-world clinical text. Text generation using clinical LLMs mitigates the risk of exposing patient privacy to improve accessing of large-scale clinical text and sharing of state-of-the-art NLP models, thus enabling the next generation clinical text analytics approaches for medical research.

Generative LLMs aspire to become a "Unified Field Theory" to unify most fundamental NLP tasks using a single model architecture. It might be still early to judge if LLMs will became the one and only foundation model[12] for NLP, but it looks like we are closer than any time. Generative LLMs have the potential to impact medical research in many aspects. In addition to performance improvement demonstrated in this study, generative LLMs provide a generalizable way for biomedical NLP using prompt-based text generation,[27] which have better few-shot learning and transfer learning ability to deliver portable clinical NLP systems. The evaluation of text generation shows that clinical LLMs can be used to generate clinical-relevant content with the potential to help document,[3] and code patient information in EHR systems, thus reducing the extensively onerous documentation burden for clinicians.[28–30] The prompt-based text generation of LLMs can potentially help compose treatment plans by integrating instructions from clinical guidelines and patient's historical records in EHRs. The conversation ability of LLMs provides opportunities developing intelligent EHR systems with human-like communication,[2] where healthcare providers, patients, and other stakeholders can communicate with electronic health record (EHR) systems in an intelligent EHR systems. Industry stakeholders such as Epic and Nuance have been reported to be exploring these potentials.[31,32]

Our Turing test focuses on (1) comparing synthetic and human notes in terms of linguistic readability and clinical relevance; and (2) testing whether physicians can differentiate synthetic and human notes. The statistical tests show that there are no significant differences in linguistic readability ($p = 0.22$; 6.57 of GatorTronGPT compared with 6.93 of human) or clinical relevance ($p = 0.91$; 7.0 of GatorTronGPT compared with 6.97 of human). Further, physicians cannot differentiate them ($p < 0.001$), suggesting the potential utility of GatorTronGPT for text generation in healthcare. Two physician evaluators find that the text written by GatorTronGPT generally lack clinical logic, indicating that more research and development are needed to make this technology useful for healthcare. Our Turing test focuses on statistical differences not utility in real-word clinical practice, which should be examined in future studies when this technology matures. Current general-purpose LLMs are designed for conversation as a chatbot outside of healthcare as there is only a small amount of biomedical text in the development dataset.

Therefore, current use of ChatGPT for healthcare is more like a typical case of intended use versus actual use as described in the medical device regulation.[33] Domain-specific LLMs are required for clinical applications. Due to the probabilistic nature of text generation, LLMs are prone to confabulation or hallucination, which might be amusing as chatbots but dangerous for healthcare. Future studies should examine strategies to control the hallucinations under a minimal level to make LLMs safe for healthcare. Like any medical AI applications, it is necessary to carefully examine potential limitations, biases, and risks of this disruptive new technology to guide its application and make it "approved" AI-enabled medical device[34] if it turns out could help healthcare. We evaluated the text generation capacity of GatorTronGPT without using human instructions, which is a typical zero-shot learning setting. Future studies should examine if the clinical text generation can be improved and controlled using human instructions such as reinforcement learning from human feedback[35] (RLFHF, used by ChatGPT) and P-tuning[36] algorithms.

**Methods**

**Data Source**

This study uses a large collection of 82 billion words of clinical narratives from UF Health Integrated Data Repository (IDR) and 195 billion words of diverse English words from the Pile[16] corpus. This study was approved by the UF Institutional Review Board (IRB202102223). At UF Health, we collected approximately 290 million clinical notes from 2011-2021 from over 126 departments, approximately 2 million patients and 50 million encounters from inpatient, outpatient, and emergency settings. The detailed patient distribution by age, gender, race, ethnicity; clinical notes distribution by note type, and clinical department can be accessed from our previous study[15]. We merged the UF Health clinical corpus with the Pile[16] dataset to generate a large corpus with 277 billion diverse clinical and English words. We performed minimal preprocessing for the Pile dataset and applied a de-identification system to remove 18 PHI categories defined in the Health Insurance Portability and Accountability Act (HIPAA) from the UF Health notes. The detailed preprocessing steps are described in the Supplement.

**Train GatorTronGPT from scratch**

**Configuration** We trained GatorTronGPT using two configurations (5 billion parameters and 20 billion parameters) and determined the number of layers, hidden sizes, and number of attention heads according to the guidelines for optimal depth-to-width parameter allocation proposed by Levin et al[37] as well as our previous experience in developing GatorTron[15]. The 5 billion model has 24 layers, hidden size of 4,096, and number of attention heads of 32; the 20 billion model has 44 layers, hidden size of 6,144, and number of attention heads of 48. We trained the 5 billion model using a 2-way tensor model parallel with a batch size of 1,120 and learning rate of 1.200E-05. We trained the 20 billion model using an 8-way tensor model parallel with a batch size of 560 and a learning rate of 1.000E-05. We adopted a dropout rate of 0.1.

**Training from scratch** We inherited the GPT-3 architecture implemented in the MegaTron-LM[38] and trained GatorTronGPT models from scratch with the default GPT-3 loss function.[13] We used a total number of 560 NVIDIA DGX A100 GPUs from 70 superPOD nodes at UF's HiPerGator-AI cluster to train GatorTronGPT by leveraging both data-level and model-level parallelisms implemented by the Megatron-LM package[38]. (See https://github.com/NVIDIA/Megatron-LM for more details) We monitored the training progress by training loss and validation loss using 3% of the data and stopped the training when there was no further improvement.

**GatorTronGPT for end-to-end biomedical relation extraction and question answering**

End-to-end relation extraction is an NLP task to identify the triplets <*concept1, concept2, relation*> from biomedical text. Question answering is to identify the *answer* for a given *question* and the *context*. Following previous studies[18,39], we approached the two tasks using a unified prompt-based text generation architecture. Specifically, we adopted a fixed-LLM prompt-tuning strategy[40] to attach a continuous embedding (i.e., virtue tokens) to the input sequence [*virtual tokens; x; y*] as a soft prompt to control the text generation; the LLM was not changed during training. We provide details in the Supplement.

**Task 1 - End-to-end biomedical relation extraction.** We compared the two GatorTronGPT models with four existing transformer models including GPT-2,[41] REBEL, REBEL-pt,[27] and BioGPT[18] on three biomedical tasks for end-to-end relation extraction using 3 benchmark datasets including drug-drug interaction[42] (DDI), BioCreative V chemical-disease relation[43] (BC5CDR), and drug-target interaction[44] (KD-DTI)

**Task 2 - Biomedical question answering.** We compared GatorTronGPT with six existing transformer models using three widely used benchmark dataset including PubMedQA[45] – a biomedical question answering dataset collected from PubMed abstracts, which requires answering questions with '*yes/no/maybe*' ; MedMCQA[46] – a large-scale multi-choice question answering dataset designed to address real world medical entrance exam questions covering 2,400 healthcare topics and 21 medical subjects; and MedQA-USMLE[47] – a multi-choice dataset collected from the professional medical board exams. These three question answering datasets have been widely used by recent studies[18,45–47] for evaluation of generative LLMs.

**Task 3 - GatorTronGPT for synthetic clinical text generation**

We sought to test the hypothesis that LLMs can generate synthetic clinical text to train synthetic NLP models useful for medical research. We applied GatorTronGPT to generate synthetic clinical text according to a set of seeds without any fine-tuning, which is a typical zero-shot learning setting. Then, using the generated synthetic clinical text, we trained synthetic transformer-based NLP models using our previous BERT-based GatorTron architecture[15], denoted as GatorTronS ('S' stands for synthetic). We trained GatorTronS models using different sizes of synthetic clinical text and compared them with the original GatorTron-base models trained using real-world text to examine how the size of synthetic clinical text affect the performance. To make it comparable, we trained GatorTronS using the same architecture and number of parameters (i.e., 345 million) as the GatorTron-base architecture. We provide detailed information in the Supplement.

**Synthetic clinical text generation**

Following previous studies[48], we approached synthetic clinical text generation as an iterative sampling procedure and applied *top-p* (i.e., nucleus sampling) sampling and temperature sampling to balance the diversity and quality of clinical text generation.[48] We set the parameter of *top-p* sampling at 0.9 and the parameter for temperature sampling at 1.2 according to our empirical assessment. We sampled the beginning 15 tokens from all sections of the de-identified notes of the MIMIC III database[49] and generated approximately 8 million prompts. We also tried several random seeds in GatorTronGPT to generate multiple documents from one prompt. We limited our clinical text generation up to 512 tokens and stopped generation when the maximum length was reached. We provide detailed information in the Supplement.

**Synthetic NLP model development**

We controlled the generation to generate different sizes of synthetic clinical text including 1 billion, 5 billion, 10 billion, and 20 billion words of clinical text and developed corresponding synthetic NLP models, denoted as GatorTronS. Following our previous study[15], we trained GatorTronS using the same architecture of GatorTron – a BERT architecture with 345 million parameters.

**Comparison with existing transformer models**

We compared GatorTronS trained using different amount of synthetic clinical text data with ClinicalBERT[50] – a clinical transformer model trained using biomedical literature and clinical notes from the MIMIC III database, and GatorTron[15], the current largest clinical transformer model trained using >90 billion words of text, using 5 clinical NLP tasks including clinical concept extraction (or named entity recognition [NER]), medical relation extraction, semantic textual similarity, natural language inference, and question answering.

**Task 4 - Turing test of text generation for clinical practice**

We randomly sampled 30 narrative sections of real-world UF Health clinical notes, including "past medical history", "history of present illness", "assessment/plan", and "chief complaint". For each of the 30 sections, we extracted the beginning 15 tokens as a seed for GatorTronGPT to generate a synthetic paragraph up to 512 tokens. We cut off the 30 real-world clinical sections to 512 tokens, removed all format information, and randomly mixed them with 30 synthetic sections written by GatorTronGPT. Two UF Health physicians (NSO, MMA) manually reviewed the 60 paragraphs of notes to evaluate: (1) linguistic readability on a 1(worst) to 9 (best) scale, (2) clinical relevance and consistency on a 1 to 9 scale, (3) determine if it was written by a human physician or GatorTronGPT. Percent agreement and Gwet's $AC_1$ were calculated to evaluate interrater reliability.[51]

## Data availability

The benchmark datasets that support the findings of this study are available from the official websites of natural language processing challenges with Data Use Agreements.

## Code Availability

The computer codes to train GatorTronGPT models are available from:

https://github.com/NVIDIA/Megatron-LM/blob/main/pretrain_gpt.py

The scripts used for data preprocessing, vocabulary training and other utilities are available from:

https://github.com/uf-hobi-informatics-lab/GatorTronGPT

The computer codes to train GatorTronS models are available from:
https://github.com/NVIDIA/Megatron-LM and https://github.com/NVIDIA/NeMo

The synthetic clinical transformer model, GatorTronS, are available from:

https://catalog.ngc.nvidia.com/orgs/nvidia/teams/clara/models/gatortron_s

The GatorTron model trained using real-world clinical text is available:

https://catalog.ngc.nvidia.com/orgs/nvidia/teams/clara/models/gatortron_og

The computer codes for preprocessing of text data are available from:

https://github.com/uf-hobi-informatics-lab/NLPreprocessing

## Acknowledgments

We would like to thank the UF Research Computing team, led by Dr. Erik Deumens, for providing computing power through UF HiPerGator-AI cluster.

## Author contributions

YW, JB, XY, NP, ABC and MGF were responsible for the overall design, development, and evaluation of this study. XY, CP, AC, and KES had full access to all the data in the study and takes responsibility for the integrity of the data and the accuracy of the data analysis. YG and YW designed the Turing evaluation of synthetic clinical text generated by GatorTronGPT. NSO and MMA are the two human physicians who performed Turing test. YW, XY, KES, CP, YG, and JB did the bulk of the writing, WH, EAS, DAM, TM, CAH, ABC, and GL also contributed to writing and editing of this manuscript. All authors reviewed the manuscript critically for scientific content, and all authors gave final approval of the manuscript for publication.

**Competing interests**

The Authors declare no Competing Financial or Non-Financial Interests.

# Supplementary Information

## Preprocessing and de-identification of clinical text

Following our previous study[1], we performed minimal preprocessing including (1) removing empty and duplicated clinical notes, unifying all text into UTF-8 encoding, and removing illegal UTF-8 strings; (2) normalizing special characters (e.g., '&', '\xa0'); (3) tokenization and sentence boundary detection. We applied a de-identification system to remove protected health information (PHI) from UF Health clinical text. (Approved under IRB202100049) We adopted the safe-harbor method to identify 18 PHI categories defined in the Health Insurance Portability and Accountability Act (HIPAA) and replaced them with dummy strings (e.g., replace people's names into [**NAME**]).

## GatorTronGPT for synthetic text generation

The goal of text generation is to generate new text content based on given text passages or prompts, which is the foundation for various large language model applications such as abstract generation and story generation. We approached the synthetic clinical text generation as an open-ended text-to-text generation task[2,3], where the generated clinical text is restricted by the context (e.g., the prompts). Specifically, given a sequence of $m$ tokens $X_{pre} = x_1 x_2 \ldots x_m$ as input context, the task is to generate the next $n$ continuation tokens $X_{cont} = x_{m+1} x_{m+2} \ldots x_{m+n}$ until reaching the max length of 512 tokens. We generate text through iteratively sampling from the pre-trained language model GatorTronGPT one token at a time by conditioning on the preceding context:

$$P(x_{cont}|x_{pre}) = \prod_{i=m+1}^{m+n} P(x_i|x_1 \ldots x_{i-1})$$

where $P(x_i|x_1 \ldots x_{i-1})$ is the next token distribution. We adopt *Top-p* (nucleus) sampling [4] during sampling to select words whose cumulative probability exceeds a predefined threshold $p$.

$$\sum_{x \in V^{(p)}} P(x|x_{1:i-1}) \geq p$$

where $V^{(p)}$ is the top-p vocabulary used to sample the next word. This approach dynamically adapts the number of words considered at each step based on their probabilities, balancing diversity and coherence of the generated text.

## GatorTronGPT for biomedical relation extraction and question answering

Following the previous study[5], we formulated both biomedical relation extraction and question answering as a prompt-based text generation model and applied prompt-tuning (p-tuning) algorithms.

**Biomedical relation extraction**. We concatenate learnable soft prompts (also called virtual prompt embeddings) with the word embeddings from the *context* (i.e., input sentence). The sample sequence is constructed as [*prompt*, *context*, *relation*], where the prompt is generated using a LSTM model and the *relation* is the gold standard label including the head entity, tail entity, and their relation type. During the inference, the *context* and the *prompt* are used as the input for our GatorTronGPT model to condition and let the model generate the relations. We converted the original relation triplets into a sequence representation. For example, there is an "*agonist*" relation between a drug - "*Igmesine*" and a target "*Opioid receptor sigma 1*", which was converted as: "the relation between [*Igmesine*] and [*Opioid receptor sigma 1*] is [*agonist*]". Thus, the relation extraction can be solved as a text generation. During inference, we converted the generated text back to triplets for evaluation. We fine-tuned and evaluated our GatorTronGPT on the end-to-end relation extraction task across four biomedical datasets: BC5CDR (chemical–disease–relation extraction), KD-DTI (drug–target–interaction extraction), DDI (drug–drug–interaction extraction) and 2018 n2c2 (Drug-ADE-relation extraction). The precision, recall, and F1 score were used for evaluation.

**Question answering.** Given a question, a context, and candidate answers, we concatenated the context and the candidate answers into a source sequence and compose the target sequence as: "the answer to the question given possible options is:", "answer": "C". Then, we adopted soft prompts instead of hard prompts (manually designed clear text phrases) in p-tuning. Specifically, we used a randomly initiated continuous embedding as soft prompts, which were fine-tuned in the training. For the PubMedQA dataset, we explored the provided artificially generated text data. Specifically, we automatically labeled the generated text using our p-tuning model developed using the training set and experimented to feedback different proportion of auto-labeled data into training. The best performance was achieved by using 5% of the auto-labeled artificially generated text data. For p-tuning, we used the implementation in NVIDIA NeMo[6], which is optimized for LLMs. We used the following parameters in our p-tuning: a global batch size of 32, virtual tokens for p-tuning 15, encoder MLP with encoder hidden size of 2,048, max sequence length of 4,096 for PubMedQA (long abstracts), 2,048 for MedMCQA and MedQA-USMLE, and a fused Adam optimizer with a learning rate of 1e-4 and a weight decay of 0.01, betas of 0.9 and 0.98, a cosine annealing scheduler monitoring validation loss with a 50 step warm up.

For example, the below is a prompt we used for MedQA-USMLE.

{"taskname": "usmle-qa", "prompt": "QUESTION: A 23-year-old man comes to the physician for evaluation of decreased hearing, dizziness, and ringing in his right ear for the past 6 months. Physical examination shows multiple soft, yellow plaques and papules on his arms, chest, and back. There is sensorineural hearing loss and weakness of facial muscles bilaterally. His gait is unsteady. An MRI of the brain shows a 3-cm mass near the right internal auditory meatus and a 2-cm mass at the left cerebellopontine angle. The abnormal cells in these masses are most likely derived from which of the following embryological structures?\nMULTIPLE CHOICES: (A) Neural tube\n(B) Surface ectoderm\n(C) Neural crest\n(D) Notochord\nTARGET: the answer to the question given possible options is: ", "answer": "C"}

## Introduction to existing transformer models for comparison

**GPT-2.** GPT-2 was trained using text data from 8 million webpages with 1.5 billion parameters, which is a scale-up of the first generation of GPT45 model. The GPT model outperformed previous transformer models on 9 out of 12 NLP tasks, whereas, the GPT-2 model further demonstrated text generation ability, which laid foundation for complex NLP tasks such as machine reading comprehension and question answering.

**REBEL and REBEL-pt.** REBEL is a transformer model based on the BART architecture designed for end-to-end relation extraction using sequence-to-sequence modeling, which outperformed previous relation extraction models based on classifications. REBEL-pt is an enhanced version of REBEL by further fine-tuning it using the triplets derived using Wikipedia hyperlinks.

**BioGPT.** BioGPT is a domain-specific generative transformer-based LLM developed using the GPT-2 architecture and the Pubmed biomedical literature, which achieved good performance in NLP tasks including relation extraction and question answering in the biomedical domain.

**Table S1. Percent agreement and interrater reliability for readability.**

|  |  | Physician 1 | | |
|---|---|---|---|---|
|  |  | High | Low | Total |
| **Physician 2** | High | 42 | 3 | 45 |
|  | Low | 10 | 5 | 15 |
|  | Total | 52 | 8 | 60 |

Percent agreement = 0.78, interrater reliability (Gwet's $AC_1$)[7] = 0.69

**Table S2. Percent agreement and interrater reliability for clinical relevance.**

|  |  | Physician 1 | | |
|---|---|---|---|---|
|  |  | High | Low | Total |
| **Physician 2** | High | 44 | 6 | 50 |
|  | Low | 7 | 3 | 10 |
|  | Total | 51 | 9 | 60 |

Percent agreement = 0.78, interrater reliability (Gwet's $AC_1$)[7] = 0.70